\documentclass{juliacon}
\usepackage{amsmath}
\usepackage{parskip}

\newtheorem{remark}{Remark}

\begin{document}

% **************GENERATED FILE, DO NOT EDIT**************

\title{RobustNeuralNetworks.jl: a Package for Machine Learning and Data-Driven Control with Certified Robustness}

\author[1]{Nicholas H. Barbara}
\author[1]{Max Revay}
\author[1]{Ruigang Wang}
\author[1]{Jing Cheng}
\author[1]{Ian R. Manchester}
\affil[1]{University of Sydney, Australian Centre for Robotics}


\hypersetup{
pdftitle = {RobustNeuralNetworks.jl: a Package for Machine Learning and Data-Driven Control with Certified Robustness},
pdfsubject = {JuliaCon 2022 Proceedings},
pdfauthor = {Nicholas H. Barbara, Max Revay, Ruigang Wang, Jing Cheng, Ian R. Manchester},
pdfkeywords = {Robustness, Machine Learning, Image Classification, Reinforcement Learning, State Estimation, Data-Driven Control},
}

\maketitle

% Abstract
\begin{abstract}

Neural networks are typically sensitive to small input perturbations, leading to unexpected or brittle behaviour. 
We present \verb|RobustNeuralNetworks.jl|: a Julia package for neural network models that are constructed to naturally satisfy a set of user-defined robustness metrics. The package is based on the recently proposed Recurrent Equilibrium Network (REN) and Lipschitz-Bounded Deep Network (LBDN) model classes, and is designed to interface directly with Julia's most widely-used machine learning package, \verb|Flux.jl|. We discuss the theory behind our model parameterization, give an overview of the package, and provide a tutorial demonstrating its use in image classification, reinforcement learning, and nonlinear state-observer design.

\end{abstract}

% Include source files
\section{Introduction} \label{sec:introduction}
Modern machine learning relies heavily on rapidly training and evaluating neural networks in problems ranging from image classification \cite{He++2016} to robotic control \cite{Siekmann++2021a}. Most neural network architectures have no robustness certificates, and can be sensitive to adversarial attacks and other input perturbations \cite{Huang++2017}. Many approaches that address this brittle behaviour rely on explicitly enforcing constraints during training to smooth or stabilize the network response \cite{Pauli++2022,Junnarkar++2023}. While effective on small-scale problems, these methods are computationally expensive, making them slow and difficult to scale up to complex real-world problems.

Recently, we proposed the \textit{Recurrent Equilibrium Network} (REN) \cite{Revay++2023} and \textit{Lipschitz-Bounded Deep Network} (LBDN) or \textit{sandwich layer} \cite{Wang+Manchester2023} model classes as computationally efficient solutions to these problems. RENs are flexible in that they include many common neural network models, such as multi-layer-perceptrons (MLPs), convolutional neural networks (CNNs), and recurrent neural networks (RNNs). Their weights and biases are parameterized to naturally satisfy a set of user-defined robustness metrics constraining the internal stability and input-output sensitivity of the network. When a network is guaranteed to satisfy a robust metric, we call this a \textit{robustness certificate}. An example is a Lipschitz bound, which restricts the network's amplification of input perturbations in its outputs \cite{Pauli++2022}. LBDNs are specializations of RENs with the specific feed-forward structure of deep neural networks like MLPs or CNNs, and built-in restrictions on the Lipschitz bound.

This special parameterization of RENs and LBDNs means that we can train models with standard, unconstrained optimization methods (such as stochastic gradient descent) while also guaranteeing their robustness. Achieving the “best of both worlds” in this way is the main advantage of the REN and LBDN model classes, and allows the user to freely train robust models for many common machine learning problems, as well as for more challenging real-world applications where safety is critical.

This papers presents \verb|RobustNeuralNetworks.jl|: a package for neural networks with built-in robustness certificates. The package contains implementations of the REN and LBDN model classes introduced in \cite{Revay++2023} and \cite{Wang+Manchester2023}, respectively, and relies heavily on key features of the Julia language \cite{Bezanson++2017} (such as multiple dispatch) for an efficient implementation of these models. The purpose of \verb|RobustNeuralNetworks.jl| is to make our recent research in robust machine learning easily accessible to users in the scientific and machine learning communities. We have therefore designed the package to interface directly with \verb|Flux.jl| \cite{Innes2018}, Julia's most widely-used machine learning package, making it straightforward to incorporate our robust neural networks into existing Julia code.

The paper is structured as follows. Section \ref{sec:overview} provides an overview of the \verb|RobustNeuralNetworks.jl| package, including a brief introduction to the model classes (Sec. \ref{sec:model-structures}), their robustness certificates (Sec. \ref{sec:robustness}), and their implementation (Sec. \ref{sec:parameterizations}). Section \ref{sec:examples} guides the reader through a tutorial with three examples to demonstrate the use of RENs and LBDNs in machine learning: image classification (Sec. \ref{sec:mnist}), reinforcement learning (Sec. \ref{sec:rl}), and nonlinear state-observer design (Sec. \ref{sec:observer}). Section \ref{sec:conc} offers some concluding remarks and future directions for robust machine learning with \verb|RobustNeuralNetworks.jl|. For more detail on the theory behind RENs and LBDNs, and for examples comparing their performance to current state-of-the-art methods on a range of problems, we refer the reader to \cite{Revay++2023} and \cite{Wang+Manchester2023} (respectively).

\section{Package overview} \label{sec:overview}
\verb|RobustNeuralNetwork.jl| contains two classes of neural network models: RENs and LBDNs. This section gives a brief overview of the two model architectures and how they are parameterized to automatically satisfy robustness metrics. We also provide some background on the different types of robustness metrics used to construct the models.

% Model types
\subsection{What are RENs and LBDNs?} \label{sec:model-structures}

A \textit{Lipschitz-Bounded Deep Network} (LBDN) is a (memoryless) deep neural network with a built-in upper-bound on its Lipschitz constant (Sec. \ref{sec:robustness-lipschitz}). Suppose the network has inputs $u \in\mathbb{R}^{n_u}$, outputs $y \in \mathbb{R}^{n_y}$, and hidden units $z_k \in \mathbb{R}^{n_k}$. The structure of an LBDN is an $L$-layer feed-forward network (like an MLP or CNN)
\begin{align} 
z_0 &= x \label{eqn:lbdn-z0}\\
z_{k+1} &= \sigma(W_k z_k + b_k), \quad k = 0, \ldots, L-1 \label{eqn:lbdn-sandwich}\\
y &= W_L z_L + b_L, \label{eqn:lbdn-output}
\end{align}
where the $W_k, b_k$ are the layer weights and biases (respectively), and $\sigma$ is a nonlinear activation function (e.g. $\tanh$, ReLU).

\begin{figure}[h]
    \centering
    \vspace{-2mm}
    \includegraphics[width=0.19\textwidth]{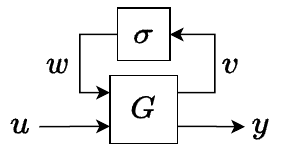}
    \vspace{-2mm}
    \caption{Feedback structure of a recurrent equilibrium network.}
    \label{fig:ren}
\end{figure}

A \textit{Recurrent Equilibrium Network} (REN) is a recurrent model (with memory) described by a linear dynamical system in feedback with a nonlinear activation function. Writing $x_t \in \mathbb{R}^{n_x}$ for the internal states of the system, a REN can be expressed mathematically as
\begin{align}
    \begin{bmatrix}
        x_{t+1} \\ v_t \\ y_t
    \end{bmatrix}&=
    \overset{W}{\overbrace{
    		\left[
    		\begin{array}{c|cc}
    		A & B_1 & B_2 \\ \hline 
    		C_{1} & D_{11} & D_{12} \\
    		C_{2} & D_{21} & D_{22}
    		\end{array} 
    		\right]
    }}
    \begin{bmatrix}
        x_t \\ w_t \\ u_t
    \end{bmatrix}+
    \overset{b}{\overbrace{
    		\begin{bmatrix}
    		b_x \\ b_v \\ b_y
    		\end{bmatrix}
    }}, \label{eqn:ren-G}\\
    w_t=\sigma(&v_t):=\begin{bmatrix}
    \sigma(v_{t}^1) & \sigma(v_{t}^2) & \cdots & \sigma(v_{t}^q)
    \end{bmatrix}^\top, \label{eqn:ren-sigma}
\end{align}
where $v_t, w_t \in \mathbb{R}^{n_v}$ are the inputs and outputs of the activation function $\sigma$. Graphically, this is equivalent to Figure \ref{fig:ren}, where the linear system $G$ is given by Equation \ref{eqn:ren-G}.

\begin{remark}
	\cite{Revay++2023} makes special mention of ``acyclic'' RENs, which have a lower-triangular $D_{11}$ matrix. Acyclic RENs are significantly more efficient to evaluate than RENs with a dense $D_{11}$ matrix, and performance is typically similar across a range of problems. All RENs in \verb|RobustNeuralNetworks.jl| are acyclic RENs. All LBDNs are acyclic by definition (see \cite[Eqn.~4]{Wang+Manchester2023}).
\end{remark}

% Robustness
\subsection{Robustness metrics and IQCs} \label{sec:robustness}

All neural network models in \verb|RobustNeuralNetworks.jl| are designed to satisfy a set of user-defined robustness metrics. We consider four particular robustness metrics relating to the internal stability of a model and its input-output map. LBDNs are more specialized and are specifically constructed to have a finite, user-tunable Lipschitz bound (Sec. \ref{sec:robustness-lipschitz}).

\subsubsection{Contracting systems} \label{sec:robustness-contraction}

\begin{figure}[ht]
    \centering
    \includegraphics[width=0.45\textwidth]{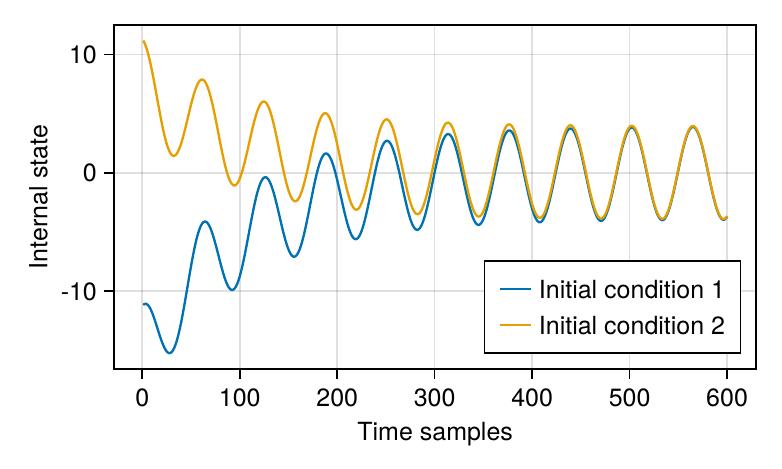}
    \vspace{-2mm}
    \caption{Simulation of a contracting REN with a single internal state. The system is simulated from two different initial states with the same sinusoidal input. The contracting system exponentially forgets its initial condition.}
    \label{fig:contracting-ren}
\end{figure}

Firstly, all of our RENs are contracting systems. This means that they exponentially ``forget'' their initial conditions. If the system starts at two different initial conditions but is given the same input sequence, the internal states will exponentially converge over time. Figure \ref{fig:contracting-ren} shows an example of a contracting REN with one input and a single internal state, where two simulations of the system start with different initial conditions but are provided the same sinusoidal input. See \cite{Lohmiller+Slotine1998} for a detailed introduction to contraction theory for dynamical systems.

\subsubsection{Incremental IQCs}

We can define additional robustness metrics for the input-output map of RENs with incremental \textit{integral quadratic constraints} (IQCs) \cite{Megretski+Rantzer1997}. Suppose we have a model $\mathcal{M}$ starting at two different initial conditions $a,b$ with two different input signals $u, v$, and consider their corresponding output trajectories $y^a = \mathcal{M}_a(u)$ and $y^b = \mathcal{M}_b(v).$ The model $\mathcal{M}$ satisfies the IQC defined by matrices $(Q, S, R)$ if
\begin{equation}
    \sum_{t=0}^T
    \begin{bmatrix}
        y^a_t - y^b_t \\ u_t - v_t
    \end{bmatrix}^\top
    \begin{bmatrix}
        Q & S^\top \\ S & R
    \end{bmatrix}
    \begin{bmatrix}
        y^a_t - y^b_t \\ u_t - v_t
    \end{bmatrix} 
    \ge -d(a,b)
    \quad \forall \, T
\end{equation}
for a function $d(a,b) \ge 0$, $d(a,a) = 0$, where $0 \succeq Q \in \mathbb{R}^{n_y\times n_y}$ is negative semi-definite, $S\in\mathbb{R}^{n_u\times n_y},$ $R=R^\top \in \mathbb{R}^{n_u\times n_u}.$ 

In general, the IQC matrices $(Q,S,R)$ can be chosen (or optimized) to meet a range of performance criteria. The following special cases are worth noting.

\subsubsection{Lipschitz bounds (smoothness)} \label{sec:robustness-lipschitz}
If $Q = -\frac{1}{\gamma}I$, $R = \gamma I$, $S = 0$ for some $\gamma \in \mathbb{R}$ with $\gamma > 0$, the model $\mathcal{M}$ satisfies a Lipschitz bound (incremental $\ell_2$-gain bound) of $\gamma$ defined by
\begin{equation}
\|\mathcal{M}_a(u) - \mathcal{M}_b(v)\|^2 \le \gamma^2 \|u - v\|^2
\end{equation}
where $\|\cdot\|$ denotes the $\ell_2$ norm. Qualitatively, the Lipschitz bound is a measure of the network's ``smoothness''. If $\gamma$ is small, then small changes to the inputs $u,v$ induce only small changes to the model output. If $\gamma$ is large (or unbounded, as in the case of, e.g., MLPs and CNNs), then the model output can change significantly even with negligible changes to the inputs. This can make the model highly sensitive to noise, adversarial attacks, and other input disturbances.

As the name suggests, all LBDN models are constructed to have a user-tunable (or learnable) Lipschitz bound.

\subsubsection{Incremental passivity} 
Passivity is a generalized notion of energy conservation from classical mechanics \cite{vanderSchaft2017}.
We have implemented two versions of incremental passivity. In each case, the network must have the same number of inputs and outputs.

\begin{enumerate}
    \item If $Q = 0, R = -2\nu I, S = I$ where $\nu \ge 0$, the model is incrementally passive (incrementally strictly input passive if $\nu > 0$). Mathematically, the following inequality holds.
    \begin{equation}
    \langle \mathcal{M}_a(u) - \mathcal{M}_b(v), u-v \rangle \ge \nu \| u-v\|^2
    \end{equation}
    \item If $Q = -2\rho I, R = 0, S = I$ where $\rho > 0$, the model is incrementally strictly output passive. Mathematically, the following inequality holds.
    \begin{equation}
    \langle \mathcal{M}_a(u) - \mathcal{M}_b(v), u-v \rangle \ge \rho \| \mathcal{M}_a(u) - \mathcal{M}_b(v)\|^2
    \end{equation}
\end{enumerate}

Passivity properties are useful in, for example, learning stable dynamical systems \cite{Cheng++2024}. The remainder of this paper will focus on contraction and Lipschitz bounds for demonstrative purposes.

% Parameterisations
\subsection{Direct and explicit parameterizations} \label{sec:parameterizations}

The key advantage of the models in \verb|RobustNeuralNetworks.jl| is that they \textit{naturally} satisfy the robustness metrics of Section \ref{sec:robustness} -- i.e., robustness is guaranteed by construction. There is no need to impose additional (possibly computationally-expensive) constraints while training a REN or an LBDN. One can simply use unconstrained optimization methods like gradient descent and be sure that the final model will satisfy the desired properties.

We achieve this by constructing the weight matrices and bias vectors in our models to automatically satisfy specific linear matrix inequalities (see \cite{Revay++2023} for details). The learnable parameters of a REN or LBDN are a set of free, unconstrained variables $\theta \in \mathbb{R}^N$. When the set of learnable parameters is exactly $\mathbb{R}^N$ like this, we call the parameterization a \textit{direct parameterization}. Equations \ref{eqn:lbdn-output} to \ref{eqn:ren-G} describe the \textit{explicit parameterizations} of RENs and LBDNs: model structures that can be called and evaluated on data. For a REN, the explicit parameters are $\bar{\theta} := [W, b]$, and for an LBDN they are $\bar{\theta} := [W_0, b_0, \ldots, W_L, b_L]$. The mapping $\theta \mapsto \bar{\theta}$ depends on the specific robustness metrics to be imposed on the explicit model. 

\subsubsection{Implementation} \label{sec:params-implementation}
RENs are defined by two abstract types in \verb|RobustNeuralNetworks.jl|. Subtypes of \verb|AbstractRENParams| hold all the information required to directly parameterize a REN satisfying some robustness metrics. For example, to initialize the direct parameters of a \textit{contracting} REN with 1 input, 10 states, 20 neurons, 1 output, and a \texttt{relu} activation function, we use the following. The direct parameters $\theta$ are stored in \verb|model_ps.direct|. 

\begin{lstlisting}[language = Julia]
using Flux, RobustNeuralNetworks

T  = Float32
nu, nx, nv, ny = 1, 10, 20, 1
model_ps = ContractingRENParams{T}(
                nu, nx, nv, ny; nl=Flux.relu)
                
println(model_ps.direct) # Access direct params
\end{lstlisting}

Subtypes of \verb|AbstractREN| represent RENs in their explicit form which can be evaluated on data. The conversion from direct to explicit parameters $\theta \mapsto \bar{\theta}$ is performed when the REN is constructed and the explicit parameters $\bar{\theta}$ are stored in \verb|model.explicit|.

\begin{lstlisting}[language = Julia]
model = REN(model_ps)    # Create explicit model
println(model.explicit)  # Access explicit params
\end{lstlisting}

\begin{figure}[ht]
    \centering
    \includegraphics[width=0.45\textwidth]{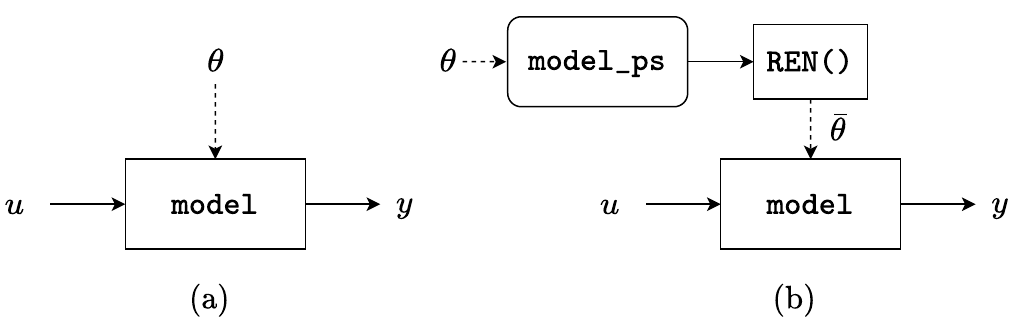}
    \caption{Association of models and their parameters in (a) \texttt{Flux.jl} and (b) \texttt{RobustNeuralNetworks.jl}. In (a), model parameters $\theta$ are associated with the \texttt{model}. In (b), the direct parameters $\theta$ are associated with the parameterization \texttt{model\_ps}, and are converted to explicit parameters $\bar{\theta}$ when the \texttt{model} is constructed for evaluation with \texttt{REN()}.}
    \label{fig:ren-params}
\end{figure}

Figure \ref{fig:ren-params} illustrates this architecture. We use a similar interface based on \verb|AbstractLBDNParams| and \verb|AbstractLBDN| for LBDNs.

% Parameterisation types
\subsubsection{Types of direct parameterizations} \label{sec:direct-params}

There are currently four REN parameterizations implemented in this package:
\begin{enumerate}
    \item \verb|ContractingRENParams| parameterizes contracting RENs with a user-defined upper bound on the contraction rate.

    \item \verb|LipschitzRENParams| parameterizes RENs with a user-defined (or learnable) Lipschitz bound $\gamma \in (0,\infty)$.

    \item \verb|PassiveRENParams| parameterizes incrementally input passive RENs with user-tunable passivity parameter $\nu \ge 0$.

    \item \verb|GeneralRENParams| parameterizes RENs satisfying some general behavioural constraints defined by an incremental IQC with parameters (Q,S,R).
    
\end{enumerate}

There is currently one LBDN parameterization implemented in \verb|RobustNeuralNetworks.jl|:

\begin{enumerate}
    \item \verb|DenseLBDNParams| parameterizes dense (fully-connected) LBDNs with a user-defined or learnable Lipschitz bound. A dense LBDN is effectively a Lipschitz-bounded MLP.
\end{enumerate}

We intend to add \verb|ConvolutionalLBDNParams| to parameterize the convolutional LBDNs in \cite{Wang+Manchester2023} in future iterations of the package.

% Explicit model wrappers
\subsubsection{Explicit model wrappers} \label{sec:explicit-wrappers}

When training a REN or LBDN, we learn and update the direct parameters $\theta$ and convert them to the explicit parameters $\bar{\theta}$ only for model evaluation. The main constructors for explicit models are \verb|REN| and \verb|LBDN|.

Users familiar with \verb|Flux.jl| will be used to creating a model once and then training it on their data. The typical workflow is as follows.

\begin{lstlisting}[language = Julia]
using Flux

# Define a model and a loss function
model = Flux.Chain(
    Flux.Dense(1 => 10, Flux.relu), 
    Flux.Dense(10 => 1, Flux.relu)
)

loss(model, x, y) = Flux.mse(model(x), y)

# Training data of 20 batches
T = Float32
xs, ys = rand(T,1,20), rand(T,1,20)
data = [(xs, ys)]

# Train the model for 50 epochs
opt_state = Flux.setup(Adam(0.01), model)
for _ in 1:50
    Flux.train!(loss, model, data, opt_state)
end
\end{lstlisting}

When training a model constructed from \verb|REN| or \verb|LBDN|, we need to back-propagate through the mapping from direct (learnable) parameters to the explicit model. We must therefore include the model construction as part of the loss function. If we do not, then the auto-differentiation engine has no knowledge of how the model parameters affect the loss, and will return zero gradients. Here is an example with an \verb|LBDN|, where the \verb|model| is defined by the direct parameterization stored in \verb|model_ps|.

\begin{lstlisting}[language = Julia]
using Flux, RobustNeuralNetworks

# Define model parameterization and loss function
T = Float32
model_ps = DenseLBDNParams{T}(1, [10], 1; nl=relu)
function loss(model_ps, x, y) 
    model = LBDN(model_ps)
    Flux.mse(model(x), y)
end

# Training data of 20 batches
xs, ys = rand(T,1,20), rand(T,1,20)
data = [(xs, ys)]

# Train the model for 50 epochs
opt_state = Flux.setup(Adam(0.01), model_ps)
for _ in 1:50
    Flux.train!(loss, model_ps, data, opt_state)
end
\end{lstlisting}

% Why separate models
\subsubsection{Separating parameters and models} \label{sec:separate-params}

For the sake of convenience, we have included the model wrappers \verb|DiffREN| and \verb|DiffLBDN| as alternatives to \verb|REN| and \verb|LBDN|, respectively. These wrappers compute the explicit parameters each time the model is called rather than just once when they are constructed. Any model created with these wrappers can therefore be used exactly the same way as a regular \verb|Flux.jl| model, and there is no need for model construction in the loss function. One can simply replace the definition of the \verb|Flux.Chain| model in the \verb|Flux.jl| example with
\begin{lstlisting}[language = Julia]
model_ps = DenseLBDNParams{T}(1, [10], 1; nl=relu)
model = DiffLBDN(model_ps)
\end{lstlisting}
and train the LBDN just like any other \verb|Flux.jl| model. We use these wrappers in Sections \ref{sec:mnist} and \ref{sec:observer}.

The trade-off in using \verb|DiffREN| or \verb|DiffLBDN| is computational efficiency in applications where a model is called many times before a training update (e.g., reinforcement learning). The main computational bottleneck in training a REN or LBDN is converting from the direct to explicit parameters (mapping $\theta \mapsto \bar{\theta}$). This process involves a matrix inverse where the number of matrix elements scales quadratically with the dimension of the model in a REN or the dimension of each layer in an LBDN (see \cite{Revay++2023,Wang+Manchester2023}). If a model is to be evaluated many times with the same direct parameters in between training updates, it is more efficient to compute the explicit parameters once, hold them fixed over many model calls, and only re-compute them once the direct parameters have been updated. This is exactly the purpose of keeping \verb|model_ps| and \verb|model| separate when using \verb|REN| and \verb|LBDN|. Note that we cannot store the direct and explicit parameters in the same \verb|model| object since auto-differentiation in Julia does not support array mutation \cite{Innes2018b}. We therefore advise using \verb|DiffREN| or \verb|DiffLBDN| for convenience in applications where the model parameters are updated after just one model call (e.g., training an image classifier). The computational benefits of separating models from their parameterizations is explored numerically in Section \ref{sec:rl}.

\section{Examples} \label{sec:examples}
This section guides the reader through a set of examples to demonstrate how to use \verb|RobustNeuralnetworks.jl| for machine learning in Julia. We will consider three examples: image classification, reinforcement learning, and nonlinear state-observer design. These examples will provide further insight into the benefits of using robust models and the reasoning behind key design decisions made in the development of the package.

We use \verb|relu| activation functions in all examples, but other choices of activation function (e.g: \verb|tanh|) are equally valid. We note that any activation function used in a REN or LBDN must have a maximum slope of 1.0, as outlined in \cite{Revay++2023,Wang+Manchester2023}. For more examples with RENs and LBDNs, please see the package documentation\footnote{\url{https://acfr.github.io/RobustNeuralNetworks.jl/}}.
\subsection{Image classification} \label{sec:mnist}

Our first example features an LBDN trained to classify the MNIST dataset \cite{LeCun++2010}. We will use this example to demonstrate how training image classifiers with LBDNs makes them robust to noise (and adversarial attacks) thanks to the built-in Lipschitz bound. For a detailed investigation of the effect of Lipschitz bounds on classification robustness and reliability, please see \cite{Wang+Manchester2023}.

% Load the data
\subsubsection{Load the data} \label{sec:mnist-data}

We begin by loading the training and test data. \verb|MLDatasets.jl|\footnote{\url{https://juliaml.github.io/MLDatasets.jl/}} contains a number of common machine-learning datasets, including the MNIST dataset. To load the full dataset of 60,000 training images and 10,000 test images, one would run the following code.

\begin{lstlisting}[language = Julia]
using MLDatasets: MNIST

T = Float32
x_train, y_train = MNIST(T, split=:train)[:]
x_test,  y_test  = MNIST(T, split=:test)[:]
\end{lstlisting}

The feature matrices \verb|x_train| and \verb|x_test| are three-dimensional arrays where each $28 \times 28$ layer contains pixel data for a single handwritten number from 0 to 9 (e.g., see Fig. \ref{fig:mnist_numbers}). The labels \verb|y_train| and \verb|y_test| are vectors containing the classification of each image as a number from 0 to 9. We convert each of these to a format better suited to training with \verb|Flux.jl|.

\begin{lstlisting}[language = Julia]
using Flux

# Reshape features for model input
x_train = Flux.flatten(x_train)
x_test  = Flux.flatten(x_test)

# Encode categorical outputs and store
y_train = Flux.onehotbatch(y_train, 0:9)
y_test  = Flux.onehotbatch(y_test,  0:9)
data = [(x_train, y_train)]
\end{lstlisting}

Features are now stored in a $28^2\times N$ \verb|Matrix| where each column contains pixel data from a single image, and the labels have been converted to a $10\times N$ \verb|Flux.OneHotMatrix| where each column contains a 1 in the row corresponding to the image's classification (e.g., row 3 for an image showing the number 2) and a 0 otherwise.

% Define a model
\subsubsection{Define a model} \label{sec:mnist-model}

We can now construct an LBDN model to train on the MNIST dataset. The larger the model, the better the classification accuracy will be, at the cost of longer training times. The smaller the Lipschitz bound $\gamma$, the more robust the model will be to input perturbations (such as noise in the image). If $\gamma$ is too small, however, it can restrict the model flexibility and limit the achievable performance \cite{Wang+Manchester2023}. For this example, we use a small network of two 64-neuron hidden layers and set a Lipschitz bound of $\gamma=5.0$ just to demonstrate the method.

\begin{lstlisting}[language = Julia]
using RobustNeuralNetworks

# Model specification
nu = 28*28              # Inputs (size of image)
ny = 10                 # Outputs (classifications)
nh = fill(64,2)         # Hidden layers 
γ  = 5.0f0              # Lipschitz bound 5.0

# Define parameters,create model
model_ps = DenseLBDNParams{T}(nu, nh, ny, γ)
model = Chain(DiffLBDN(model_ps), Flux.softmax)
\end{lstlisting}

The \verb|model| consists of two parts. The first is a callable \verb|DiffLBDN| model constructed from its direct parameterization, which is defined by an instance of \verb|DenseLBDNParams| as per Section \ref{sec:parameterizations}. The output is then converted to a probability distribution using a \verb|softmax| layer. Note that all \verb|AbstractLBDN| models can be combined with traditional neural network layers using \verb|Flux.Chain|. 

We could also construct the \verb|model| as a chain of \verb|SandwichFC| layers. Introduced in \cite{Wang+Manchester2023}, the ``sandwich'' layer is a dense layer with a guaranteed Lipschitz bound of 1.0. We have designed the user interface for \verb|SandwichFC| similarly to that of \verb|Flux.Dense|.
\begin{lstlisting}[language = Julia]
model = Chain(
    (x) -> (sqrt(γ) * x),
    SandwichFC(nu => nh[1], relu; T),
    SandwichFC(nh[1] => nh[2], relu; T),
    (x) -> (sqrt(γ) * x),
    SandwichFC(nh[2] => ny; output_layer=true, T),
    Flux.softmax
)
\end{lstlisting}
This model is equivalent to a dense LBDN constructed with \verb|LBDN| or \verb|DiffLBDN|. We have included it as a convenience for users familiar with layer-wise network construction in \verb|Flux.jl|, and recommend using it interchangeably with \verb|DiffLBDN|.

% Define a loss function
\subsubsection{Define a loss function} \label{sec:mnist-loss}

A typical loss function for training on datasets with discrete labels is the cross entropy loss. We can use the \verb|crossentropy| loss function shipped with \verb|Flux.jl|.

\begin{lstlisting}[language = Julia]
loss(model,x,y) = Flux.crossentropy(model(x), y)
\end{lstlisting}

% Train the model
\subsubsection{Train the model} \label{sec:mnist-train}

We train the model over 600 epochs using two learning rates: \verb|1e-3| for the first 300, and \verb|1e-4| for the last 300. We use the \verb|Adam| optimizer \cite{Kingma+Ba2015} and the default \verb|Flux.train!| method for convenience. Note that the \verb|Flux.train!| method updates the learnable parameters each time the model is evaluated on a batch of data, hence our choice of \verb|DiffLBDN| as a model wrapper.

\begin{lstlisting}[language = Julia]
# Hyperparameters
epochs = 300
lrs = [1e-3,1e-4]

# Train with the Adam optimizer
opt_state = Flux.setup(Adam(lrs[1]), model)
for k in eachindex(lrs)
    for i in 1:epochs
        Flux.train!(loss, model, data, opt_state)
    end
    Flux.adjust!(opt_state, lrs[2])
end
\end{lstlisting}

% Evaluate the trained model
\subsubsection{Evaluate the trained model} \label{sec:mnist-evaluate}

\begin{figure}[!t]
    \centering
    \includegraphics[width=0.47\textwidth]{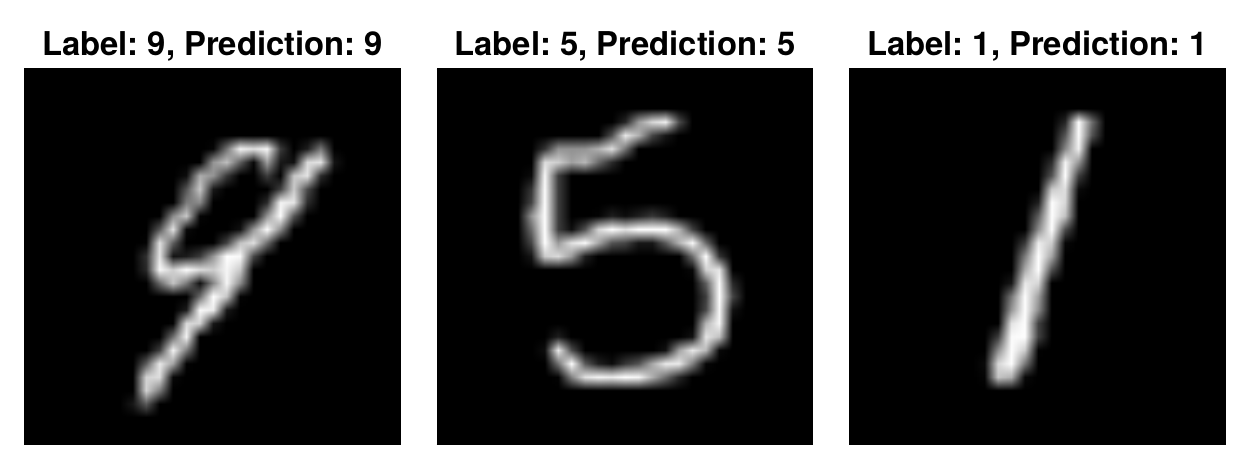}
    \caption{Examples of classifications from the trained LBDN model on the MNIST dataset.}
    \label{fig:mnist_numbers}
\end{figure}

Our final model achieves training and test accuracies of approximately 98\% and 97\%, respectively, as shown in Table \ref{tab:mnist-results}. We could improve this further by switching to a convolutional LBDN, as in \cite{Wang+Manchester2023}. Some examples of classifications given by the trained LBDN model are presented in Figure \ref{fig:mnist_numbers}.

% Investigate robustness
\subsubsection{Investigate robustness} \label{sec:mnist-robustness}

The main advantage of using an LBDN for image classification is its built-in robustness to noise (or attacks) added to the image. This robustness is a direct benefit of the Lipschitz bound. As outlined in the Section \ref{sec:robustness-lipschitz}, the Lipschitz bound effectively defines how ``smooth'' the network is: the smaller the Lipschitz bound, the less the network outputs will change as the inputs vary. For example, small amounts of noise added to the image will be less likely to change its classification. A detailed investigation into this effect is presented in \cite{Wang+Manchester2023}.

We can demonstrate the robustness of LBDNs by comparing the model to a standard MLP built from \verb|Flux.Dense| layers. We first create a \verb|dense| network with the same layer structure as the LBDN.
\begin{lstlisting}[language = Julia]
# Initialisation functions
init = Flux.glorot_normal
initb(n) = Flux.glorot_normal(n)

# Build a dense model
dense = Chain(
    Dense(nu => nh[1], relu; 
          init, bias=initb(nh[1])),
    Dense(nh[1] => nh[2], relu; 
          init, bias=initb(nh[2])),
    Dense(nh[2] => ny; init, bias=initb(ny)),
    Flux.softmax
)
\end{lstlisting}

Training the \verb|dense| model with the same training loop used for the LBDN model results in a model that achieves training and test accuracies of approximately 98\% and 97\%, respectively, as shown in Table \ref{tab:mnist-results}.

\begin{table} [ht]
\tbl{Training and test accuracy for the LBDN and Dense models on the MNIST dataset without perturbations.}{
\begin{tabular}{|l|c|c|}\hline
    \textbf{Model structure} & \textbf{Training accuracy (\%)} & \textbf{Test accuracy (\%)} \\ \hline
    LBDN & 98.2 & 97.2 \\ \hline
    Dense & 97.6 & 96.6 \\ \hline
\end{tabular}}
\label{tab:mnist-results}
\end{table}

As a simple test of robustness, we add uniformly-sampled random noise in the range $[-\epsilon, \epsilon]$ to the pixel data in the test dataset for a range of noise magnitudes $\epsilon \in [0, 200/255].$ We record the test accuracy for each perturbation size and store it for plotting.
\begin{lstlisting}[language = Julia]
using Statistics

# Get test accuracy as we add noise
uniform(x) = 2*rand(T, size(x)...) .- 1
compare(y, yh) = 
    maximum(yh, dims=1) .== maximum(y.*yh, dims=1)
accuracy(model, x, y) = mean(compare(y, model(x)))
    
function noisy_test_error(model, ϵ)
    noisy_xtest = x_test .+ ϵ*uniform(x_test)
    accuracy(model, noisy_xtest,  y_test)*100
end

ϵs = T.(LinRange(0, 200, 10)) ./ 255
lbdn_error  = noisy_test_error.((model,), ϵs)
dense_error = noisy_test_error.((dense,), ϵs)
\end{lstlisting}

\begin{figure}[!b]
    \centering
    \includegraphics[width=0.47\textwidth]{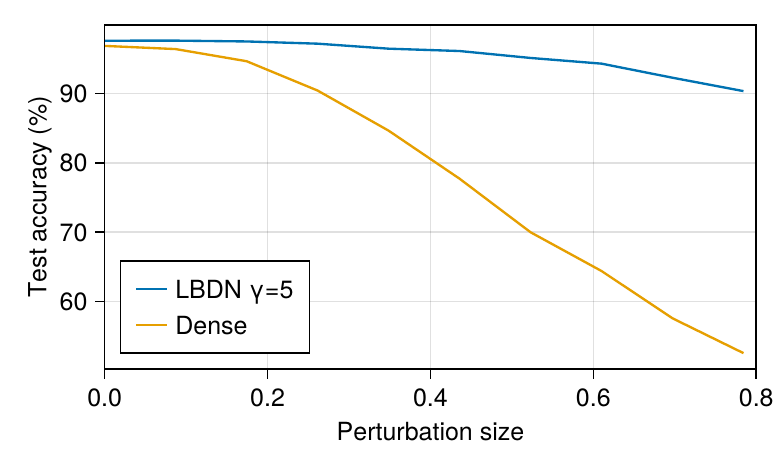}
    \caption{Comparison of test accuracy on the MNIST dataset as a function of random perturbation magnitude $\epsilon$. The LBDN model is significantly more robust than a standard \texttt{Dense} network.}
    \label{fig:mnist_robust}
\end{figure}

Plotting the results in Figure \ref{fig:mnist_robust} very clearly shows that the \verb|dense| network, which has no guarantees on its Lipschitz bound, quickly loses its accuracy as small amounts of noise are added to the image. In contrast, the LBDN \verb|model| maintains its accuracy even when the (maximum) perturbation size is as much as 80\% of the maximum pixel values. This is an illustration of why image classification is such a promising use-case for LBDN models. For a more detailed comparison of LBDN with state-of-the-art image classification methods, see \cite{Wang+Manchester2023}.

\subsection{Reinforcement learning} \label{sec:rl}

One of the original motivations for developing the model structures in  \verb|RobustNeuralNetworks.jl| was to guarantee stability and robustness in learning-based control. Recently, we have shown that with a controller architecture based on a nonlinear version of classical Youla-Ku\v{c}era parameterization \cite{Kucera1975,Youla++1976}, one can learn over a space of stabilizing controllers for linear and nonlinear systems using standard reinforcement learning techniques, so long as the control policy is parameterized by a contracting, Lipschitz-bounded REN  \cite{Wang+Manchester2022,Wang++2022,Barbara++2023}. This is an exciting result for learning-based controllers in safety-critical systems, such as in robotics.

In this example, we will demonstrate how to train an LBDN controller with \textit{reinforcement learning} (RL) for a simple nonlinear dynamical system. This controller will not have any stability guarantees. The purpose of this example is simply to showcase the steps required to set up RL experiments for more complex systems with RENs and LBDNs.

% Overview
\subsubsection{Overview} \label{sec:rl-overview}

\begin{figure}[!t]
    \centering
    \includegraphics[width=0.27\textwidth]{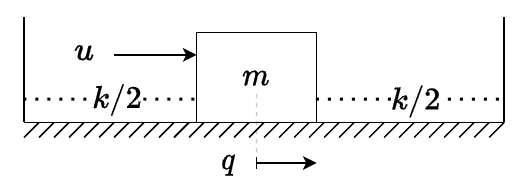}
    \caption{Mechanical system to be controlled. A box sits in a tub of fluid, suspended between two springs, and can be pushed by a force $u$ to different horizontal positions $q$.}
    \label{fig:rl-box}
\end{figure}

Consider the simple mechanical system shown in Figure \ref{fig:rl-box}: a box of mass $m$ sits in a tub of fluid, held between the walls by two springs each with spring constant $k/2.$ The box can be pushed with a force $u.$ Its dynamics are
\begin{equation} \label{eqn:box-dynamics}
m\ddot{q} = u - kq - \mu \dot{q}|\dot{q}|
\end{equation}
where $\mu$ is the viscous damping coefficient due to the box moving through the fluid, and $\dot{q},\ddot{q}$ denote the velocity and acceleration of the box, respectively.

We can write this as a (nonlinear) state-space model with state $x = (q,\dot{q})^\top,$ control input $u,$ and dynamics
\begin{equation}
\dot{x} = f(x,u) := \begin{bmatrix}
\dot{q} \\ (u - kq - \mu \dot{q}|\dot{q}|)/m
\end{bmatrix}.
\end{equation}
This is a continuous-time model of the dynamics. For our purposes, we need a discrete-time model. We can discretize the dynamics using a forward Euler approximation to get
\begin{equation}
x_{t+1} = f_d(x_t,u_t) := x_t + \Delta t \cdot f(x_t, u_t)
\end{equation}
where $\Delta t$ is the time-step. This approximation typically requires a small time-step for numerical stability, but is sufficient for our simple example. If physical accuracy was of concern, one could use a fourth (or higher) order Runge-Kutta scheme.

Our aim is to learn a controller $u = \mathcal{K}_\theta(x, q_\mathrm{ref}),$ defined by some learnable parameters $\theta,$ that can push the box to any goal position $q_\mathrm{ref}$ that we choose. Specifically, we want the box to:
\begin{enumerate}
    \item reach a (stationary) goal position $q_\mathrm{ref}$
    \item within a time period $T$.
\end{enumerate}
The force required to keep the box at a static equilibrium position $q_\mathrm{ref}$ is $u_\mathrm{ref} = k q_\mathrm{ref}$ from Equation \ref{eqn:box-dynamics}. We can encode these objectives into a cost function $J_\theta$ and write our RL problem as
\begin{equation} \label{eqn:rl-costfunc}
\min_\theta \mathbb{E} \left[ J_\theta \right],
\quad
J_\theta = \sum_{t=0}^{T-1} c_1 (\Delta q_t)^2 + c_2 \dot{q}_t^2 + c_3 (\Delta u_t)^2
\end{equation}
where $\Delta q_t = q_t - q_\mathrm{ref}$, $\Delta u_t = u_t - u_\mathrm{ref}$, $c_1, c_2, c_3$ are cost function weights, and the expectation is over different initial and goal positions of the box.

% Problem setup
\subsubsection{Problem setup} \label{sec:rl-setup}

We start by defining the properties of our system and translating the dynamics into Julia code. For this example, we consider a box of mass $m=1$, spring constants $k=5,$ and a viscous damping coefficient $\mu = 0.5$. We will simulate the system over $T = 4$\,s time horizons with a time-step of $\Delta t = 0.02$\,s.

\begin{lstlisting}[language = Julia]
m = 1                   # Mass (kg)
k = 5                   # Spring constant (N/m)
μ = 0.5                 # Viscous damping (kg/m)
Tmax = 4                # Simulation horizon (s)
dt = 0.02               # Time step (s)
ts = 1:Int(Tmax/dt)     # Array of time indices
\end{lstlisting}

Now we can generate the training data. Suppose the box always starts at rest from the zero position, and the goal position can be anywhere in the range $q_\mathrm{ref} \in [-1,1]$. Our training data consists of a batch of 80 randomly-sampled goal positions and corresponding reference forces $u_\mathrm{ref}$.
\begin{lstlisting}[language = Julia]
nx, nref, batches = 2, 1, 80
x0 = zeros(nx, batches)
qref = 2*rand(nref, batches) .- 1
uref = k*qref
\end{lstlisting}

It is good practice (and faster) to simulate all simulation batches at once, so we define our dynamics functions to operate on batches of states and controls. Each row corresponds to a different state or control, and each column corresponds to a simulation for a particular goal position.

\begin{lstlisting}[language = Julia]
f(x::Matrix,u::Matrix) = [x[2:2,:]; (u[1:1,:] - 
    k*x[1:1,:] - μ*x[2:2,:]*abs.(x[2:2,:]))/m]
fd(x::Matrix,u::Matrix) = x + dt*f(x,u)
\end{lstlisting}

RL problems typically involve simulating the system over some time horizon and collecting rewards or costs at each time step. Control policies are trained using approximations of the cost gradient $\nabla J_\theta$, as it is often difficult (or impossible) to compute the exact gradient due to the complexity of dynamics simulators. We refer the reader to \cite{Sutton+Barto2018} for further details, and \verb|ReinforcementLearning.jl| \cite{Tian++2020} for examples in Julia.

For this simple example, we can back-propagate directly through the dynamics function \verb|fd(x,u)| rather than approximating  $\nabla J_\theta$. The simulator below takes a batch of initial states, goal positions, and a controller \verb|model| whose inputs are $[x; q_\mathrm{ref}]$. It computes trajectories of states and controls $z = \{[x_0;u_0], \ldots, [x_{T-1};u_{T-1}]\}$. To avoid the issue of unsupported array mutation when differentiating we use a \verb|Zygote.Buffer| to iteratively store the outputs \cite{Innes2018b}.

\begin{lstlisting}[language = Julia]
using Zygote: Buffer

function rollout(model, x0, qref)
    z = Buffer([zero([x0;qref])], length(ts))
    x = x0
    for t in ts
        u = model([x;qref])
        z[t] = vcat(x,u)
        x = fd(x,u)
    end
    return copy(z)
end
\end{lstlisting}

After computing these trajectories, we will need a function to evaluate the cost given some weightings $c_1,c_2,c_3$.
\begin{lstlisting}[language = Julia]
using Statistics

weights = [10,1,0.1]
function _cost(z, qref, uref)
    Δz = z .- [qref; zero(qref); uref]
    return mean(sum(weights .* Δz.^2; dims=1))
end
cost(z::AbstractVector, qref, uref) = 
    mean(_cost.(z, (qref,), (uref,)))
\end{lstlisting}

% Define a model
\subsubsection{Define a model} \label{sec:rl-model}

We will train an LBDN controller with a Lipschitz bound of $\gamma = 20$. Its inputs are the state $x_t$ and goal position $q_\mathrm{ref}$, while its outputs are the control force $u_t$. We have chosen a model with two hidden layers each of 32 neurons just as an example. For examples of how Lipschitz bounds can be useful in learning robust controllers, see \cite{Barbara++2024a,Russo+Proutiere2021}.

\begin{lstlisting}[language = Julia]
using RobustNeuralNetworks

T  = Float64
γ  = 20                 # Lipschitz bound
nu = nx + nref          # Inputs (x and reference)
ny = 1                  # Outputs (control action)
nh = fill(32, 2)        # Hidden layers
model_ps = DenseLBDNParams{T}(nu, nh, ny, γ)
\end{lstlisting}

% Define a loss function
\subsubsection{Define a loss function} \label{sec:rl-loss}

In constructing a loss function for this problem, we refer to Section \ref{sec:explicit-wrappers}. The \verb|model_ps| contain all information required to define a dense LBDN model. However, \verb|model_ps| is not a model that can be evaluated on data: it is a \textit{model parameterization}, and contains the learnable parameters $\theta$. To train an LBDN given some data, we construct the model within the loss function using the \verb|LBDN| wrapper so that the mapping from direct to explicit parameters is captured during back-propagation. Our loss function therefore includes the following three components.

\begin{lstlisting}[language = Julia]
function loss(model_ps, x0, qref, uref)
    model = LBDN(model_ps)            # Model
    z = rollout(model, x0, qref)      # Simulation
    return cost(z, qref, uref)        # Cost
end
\end{lstlisting}

% Train the model
\subsubsection{Train the model} \label{sec:rl-train}

Having set up the RL problem, all that remains is to train the controller. The function below trains a model and keeps track of the training loss \verb|tloss| (cost $J_\theta$) for each simulation in our batch of 80. Training is performed with the \verb|Adam| optimizer over 250 epochs with a learning rate of $10^{-3}$.

\begin{lstlisting}[language = Julia]
using Flux

function train_box_ctrl!(
    model_ps, loss_func; 
    epochs=250, lr=1e-3
)
    costs = Vector{Float64}()
    opt_state = Flux.setup(Adam(lr), model_ps)
    for k in 1:epochs

        tloss, dJ = Flux.withgradient(
            loss_func, model_ps, x0, qref, uref)
        Flux.update!(opt_state, model_ps, dJ[1])
        push!(costs, tloss)
    end
    return costs
end

costs = train_box_ctrl!(model_ps, loss)
\end{lstlisting}

% Evaluate the trained model
\subsubsection{Evaluate the trained model} \label{sec:rl-evaluate}

\begin{figure}
    \centering
    \includegraphics[width=0.47\textwidth]{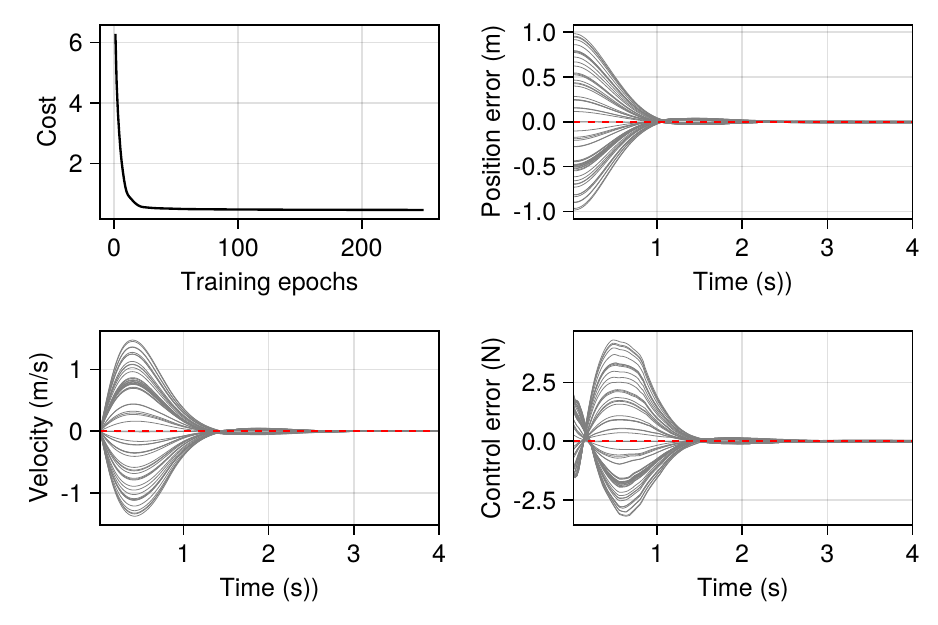}
    \caption{Loss curve and simulation results from the LBDN RL policy controlling the box system in Figure \ref{fig:rl-box}. The LBDN policy can push the box to any desired location in the domain of interest. The position and controller errors are $\Delta q$ and $\Delta u$ from Equation \ref{eqn:rl-costfunc}, respectively.}
    \label{fig:rl-results}
\end{figure}

We may now verify the performance of the trained model on a new set of reference positions. In the code below, we generate 60 batches of test data. In each one, the box starts at the origin at rest, and is moved through the fluid to a different (random) goal position $q_\mathrm{ref} \in [-1,1].$ We plot the states and controls alongside the loss curve from training in Figure \ref{fig:rl-results}. The box clearly moves to the required position within the time frame in all cases, experimentally verifying the performance of our controller.

\begin{lstlisting}[language=Julia]
model   = LBDN(model_ps)
x0_test = zeros(2,60)
qr_test = 2*rand(1, 60) .- 1
z_test  = rollout(model, x0_test, qr_test)
\end{lstlisting}

% Using DiffLBDN
\subsubsection{Advantages of separate parameters and models} \label{sec:rl-comptime}

As discussed in Section \ref{sec:separate-params}, there is a trade-off between convenience and performance in \verb|RobustNeuralNetworks.jl|. The \verb|DiffLBDN| and \verb|DiffREN| wrappers exist to allow users to train robust models in a \verb|Flux.jl|-like manner. These wrappers convert a model parameterization to its explicit form each time they are called, hence the user does \textit{not} have to re-construct the model in the loss function.
\begin{lstlisting}[language = Julia]
loss2(model, x0, qref, uref) = 
    cost(rollout(model, x0, qref), qref, uref)
\end{lstlisting}

The cost is computation speed, particular in an RL context. Careful inspection of the \verb|rollout()| function shows that the \verb|model| is evaluated many times within the loss function before the learnable parameters are updated with \verb|Flux.update!()|. As discussed in the Section \ref{sec:separate-params}, the major computational bottleneck in training RENs and LBDNs is the conversion from learnable (direct) parameters to an explicit model. Constructing the model only when the parameters are updated therefore saves considerably on computation time, particularly for large models.

For example, suppose we train single-hidden-layer LBDNs with $n = 2, 2^2, \ldots, 2^9$ neurons over 100 epochs on our box RL problem, and log the time taken to train each model when using both \verb|LBDN| and \verb|DiffLBDN|.

\newpage
\begin{lstlisting}[language = Julia]
function lbdn_compute_times(n; epochs=100)

    # Build model params and a model
    lbdn_ps = DenseLBDNParams{T}(nu, [n], ny, γ)
    diff_lbdn = DiffLBDN(deepcopy(lbdn_ps))

    # Time with LBDN vs DiffLBDN (respectively)
    t_lbdn = @elapsed (
        train_box_ctrl!(lbdn_ps, loss; epochs))
    t_diff_lbdn = @elapsed (
        train_box_ctrl!(diff_lbdn, loss2; epochs))
    return [t_lbdn, t_diff_lbdn]

end

# Evaluate computation time
# Run it once first for just-in-time compiler
ns = 2 .^ (1:9)
lbdn_compute_times(2; epochs=1)
comp_times = reduce(hcat, lbdn_compute_times.(ns))
\end{lstlisting}

\begin{figure}[ht]
    \centering
    \includegraphics[width=0.47\textwidth]{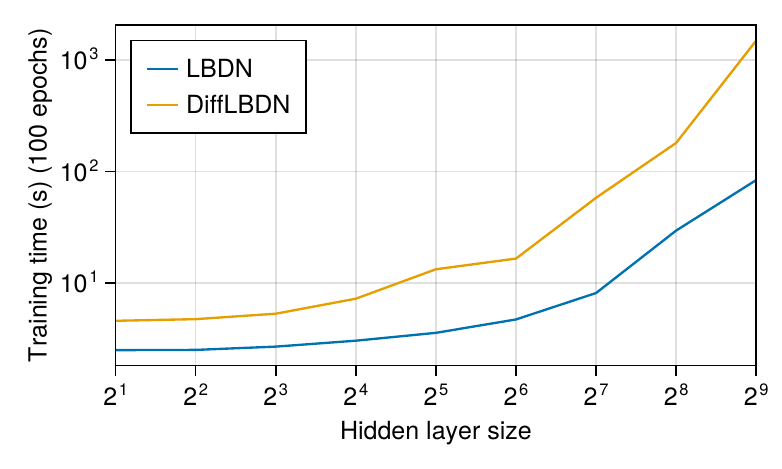}
    \caption{Training time as a function of hidden-layer size for a single-hidden-layer LBDN constructed with both the \texttt{LBDN} and \texttt{DiffLBDN} wrappers. Using the \texttt{LBDN} wrapper for RL is significantly more efficient than re-constructing the explicit model at every evaluation of the \texttt{DiffLBDN} model.}
    \label{fig:rl-comptime}
\end{figure}

The results are plotted in Figure \ref{fig:rl-comptime}. Even for a single-layer LBDN with $2^9 = 512$ neurons, it is clear that using \verb|DiffLBDN| takes an order of magnitude longer to train than only constructing the \verb|LBDN| model each time the \verb|loss()| function is called. If we were training dynamic models with REN, the computational overhead of using \verb|DiffREN| instead of \verb|REN| would be even more extreme, since the conversion from direct to explicit parameters in a REN is typically more computationally expensive than for LBDNs. It is for this reason that we strongly recommend using the \verb|LBDN| and \verb|REN| wrappers if many evaluations of the model are required before \verb|Flux.update!()| (or equivalent) is called, as in RL.

\subsection{Observer design} \label{sec:observer}

In Section \ref{sec:rl}, we designed a controller for a simple nonlinear system assuming that the controller had \textit{full state knowledge}: that is, it had access to both the position and velocity of the box. In many practical situations, we may only be able to measure some of the system states. For example, our box may have a camera to estimate its position but not its velocity. In these cases, we need a \textit{state observer} to estimate the full state of the system for feedback control.

In this example, we will show how a contracting REN can be used to learn stable observers for dynamical systems. A common approach to designing state estimators for nonlinear systems is the \textit{Extended Kalman Filter} (EKF). In our case, we will consider observer design as a supervised learning problem. For a detailed explanation of the theory behind learning state observers, and for a similar example designing an observer for a \textit{Partial Differential Equations} (PDE), please refer to Section VIII of \cite{Revay++2023}.

% Overview
\subsubsection{Background theory} \label{sec:observer-theory}

We briefly summarise some background theory from \cite{Revay++2023} relevant to this example. Suppose we have a discrete-time, nonlinear dynamical system of the form
\begin{align}
    x_{t+1} &= f_d(x_t, u_t) \\
    y_t &= g_d(x_t, u_t)
\end{align}
with state vector $x_t,$ controlled inputs $u_t,$ and measured outputs $y_t.$ Our aim is to estimate the sequence $\{x_0, x_1, \ldots, x_T \}$ over some time period $[0,T]$ given only the measurements $y_t$ and inputs $u_t$ at each time step. We will use a very general form for an observer
\begin{equation}
    \hat{x}_{t+1} = f_o(\hat{x}_t, u_t, y_t)
\end{equation}
where $\hat{x}_t$ is the state estimate. A more common (but more restrictive) structure is the well-known Luenberger observer \cite{Luenberger1971}.

To estimate the true state, our observer error $(x_t - \hat{x}_t)$ must converge to zero as time progresses, or $\hat{x}_t \rightarrow x_t$ as $t \rightarrow \infty$. As outlined in \cite{Revay++2023}, our observer only has to satisfy the following two conditions to guarantee this.

\begin{enumerate}
    \item The observer must be a contracting system (Sec. \ref{sec:robustness-contraction}).
    \item The observer must satisfy a ``correctness'' condition which says that, given perfect knowledge of the state, measurements, and inputs, the observer can exactly predict the next state. Mathematically, we write this as
    \begin{equation}
        f_o(x_t,u_t,y_t) = f_d(x_t,u_t)
    \end{equation}
    where $y_t = g_d(x_t,u_t)$. Note the use of $x_t$ not $\hat{x}_t$. It turns out that if the correctness condition is only approximately satisfied such that $|f_o(x_t,u_t,y_t) - f_d(x_t,u_t)| < \rho$ for some small number $\rho\in\mathbb{R}$, then the observer error will still be bounded. See Appendix E of \cite{Revay++2023} for details.
\end{enumerate}

The first condition, contraction, is already guaranteed for all REN models in \verb|RobustNeuralNetworks.jl|. Therefore, to learn a stable observer with RENs, our only requirement is to minimise the one-step-ahead prediction error to approximate the correctness condition. If we have a batch of data $z = \{x_i, u_i, y_i, \ i = 1,2,\ldots,N\},$ this corresponds to minimising the loss function
\begin{equation}
\mathcal{L}(z, \theta) = \sum_{i=1}^N |f_o(x_i,u_i,y_i) - f_d(x_i,u_i)|^2,
\end{equation}
where $\theta$ contains the learnable parameters of the REN.

% Problem setup
\subsubsection{Generate training data} \label{sec:observer-setup}

Consider the same nonlinear box system from Section \ref{sec:rl}, with the a change in setup so that we can only measure the box position. We introduce a measurement function \verb|gd()| such that $y_t = x_t$.

\begin{lstlisting}[language = Julia]
m = 1                   # Mass (kg)
k = 5                   # Spring constant (N/m)
μ = 0.5                 # Viscous damping (kg/m)
nx = 2                  # Number of states

f(x::Matrix,u::Matrix) = [x[2:2,:]; (u[1:1,:] - 
    k*x[1:1,:] - μ*x[2:2,:]*abs.(x[2:2,:]))/m]

fd(x,u) = x + dt*f(x,u)
gd(x::Matrix) = x[1:1,:]
\end{lstlisting}

For this example, we assume that the box always starts at rest in a random initial position between $\pm0.5$m, after which it is released and allowed to oscillate freely with no added forces (so $u = 0$). Learning an observer typically requires a large amount of training data to fully capture the behaviour of the system, hence we consider 200 batches each simulating 10\,s of motion.
\begin{lstlisting}[language = Julia]
Tmax = 10               # Simulation horizon
dt = 0.01               # Time step (s)
ts = 1:Int(Tmax/dt)     # Time array indices

# Generate batches of training data
batches = 200
u = fill(zeros(1, batches), length(ts)-1)
X = fill(zeros(1, batches), length(ts))
X[1] = 0.5*(2*rand(nx, batches) .- 1)

for t in ts[1:end-1]
    X[t+1] = fd(X[t],u[t])
end
\end{lstlisting}
We have stored the states of the system across each batch in \verb|X|. To compute the one-step-ahead loss $\mathcal{L},$ we will need to separate this data into the states at the ``current'' time step \verb|Xt| and at the ``next'' time step \verb|Xn|, then compute the measurement outputs. We then store the data for training, shuffling it so there is no bias in the training towards earlier time steps.
\begin{lstlisting}[language = Julia]
using Random

# Current/next state, measurements
Xt = X[1:end-1]
Xn = X[2:end]
y  = gd.(Xt)

# Store training data
obsv_data = [[ut; yt] for (ut,yt) in zip(u, y)]
indx = shuffle(1:length(obsv_data))
data = zip(Xn[indx], Xt[indx], obsv_data[indx])
\end{lstlisting}

% Define a model
\subsubsection{Define a model} \label{sec:observer-model}

We can construct the parameterization for a contracting REN model using \verb|ContractingRENParams|. The inputs to the model are $[u_t;y_t]$, and its outputs are the next state estimate $\hat{x}_{t+1}$. The flag \verb|output_map=false| sets the output map of the REN to just return its own internal state -- i.e., $C_2 = I$, $D_{21} = 0$, $D_{22} = 0$, $b_y = 0$ from Equation \ref{eqn:ren-G}. This makes the internal state of the REN exactly the state estimate $\hat{x}_t$.

\begin{lstlisting}[language = Julia]
using RobustNeuralNetworks

T  = Float32
nv = 200
nu = size(obsv_data[1], 1)
ny = nx
model_ps = ContractingRENParams{T}(
    nu, nx, nv, ny; output_map=false)
model = DiffREN(model_ps)
\end{lstlisting}

% Define a loss function
\subsubsection{Define a loss function} \label{sec:observer-loss}

As outlined in Section \ref{sec:observer-theory}, our loss function should be the one-step-ahead prediction error of the REN observer. We write this as follows, noting that all subtypes of \verb|AbstractREN| return both their updated internal state and their output (in that order).
\begin{lstlisting}[language = Julia]
using Statistics

function loss(model, xn, xt, inputs)
    xpred = model(xt, inputs)[1]
    return mean(sum((xn - xpred).^2, dims=1))
end
\end{lstlisting}

% Train the model
\subsubsection{Train the model} \label{sec:observer-train}

The function below trains the observer with the \verb|Adam| optimizer over 100 epochs and decreases the maximum learning rate from $10^{-3}$ to $10^{-4}$ if the mean loss stops decreasing between epochs. The core of this function is a simple \verb|Flux.jl| training loop, expanded out for clarity.
\begin{lstlisting}[language = Julia]
using Flux

function train_observer!(
    model, data; 
    epochs=50, lr=1e-3, min_lr=1e-6
)
    opt_state = Flux.setup(Adam(lr), model)
    mean_loss = [1e5]
    for epoch in 1:epochs

        # Gradient descent update
        batch_loss = []
        for (xn, xt, inputs) in data
            tloss, dJ = Flux.withgradient(
                loss, model, xn, xt, inputs)
            Flux.update!(opt_state, model, dJ[1])
            push!(batch_loss, tloss)
        end

        # Reduce lr if loss is stuck or growing
        push!(mean_loss, mean(batch_loss))
        if (mean_loss[end] >= mean_loss[end-1]) && 
           (lr > min_lr)
            lr *= 0.1
            Flux.adjust!(opt_state, lr)
        end
    end
    return mean_loss
end
tloss = train_observer!(model, data)
\end{lstlisting}

% Evaluate the trained model
\subsubsection{Evaluate the trained model} \label{sec:observer-evaluate}

We have trained the REN observer to minimise the one-step-ahead prediction error, but we are yet to test whether the the observer error actually does converge to zero. We set up the following 50 batches of test data as a demonstration.
\begin{lstlisting}[language = Julia]
batches   = 50
ts_test   = 1:Int(20/dt)
u_test    = fill(zeros(1, batches),length(ts_test))
x_test    = fill(zeros(nx,batches),length(ts_test))
x_test[1] = 0.2*(2*rand(nx, batches) .-1)

for t in ts_test[1:end-1]
    x_test[t+1] = fd(x_test[t], u_test[t])
end
y_test = gd.(x_test)
obsv_in = [[u;y] for (u,y) in zip(u_test, y_test)]
\end{lstlisting}

Next, we need a function to simulate the REN observer using its own state $\hat{x}_t$ rather than the true system state $x_t$, which was used for training. We use the very neat tool \verb|Flux.Recur| for this. We assume that the observer has no knowledge of the initial state and simply guesses $\hat{x}_0 = 0$ for all 50 batches.
\begin{lstlisting}[language = Julia]
function simulate(model::AbstractREN, x0, u)
    recurrent = Flux.Recur(model, x0)
    output = recurrent.(u)
    return output
end
x0hat = zeros(model.nx, batches)
xhat = simulate(model, x0hat, obsv_in)
\end{lstlisting}

\begin{figure}
    \centering
    \includegraphics[width=0.47\textwidth]{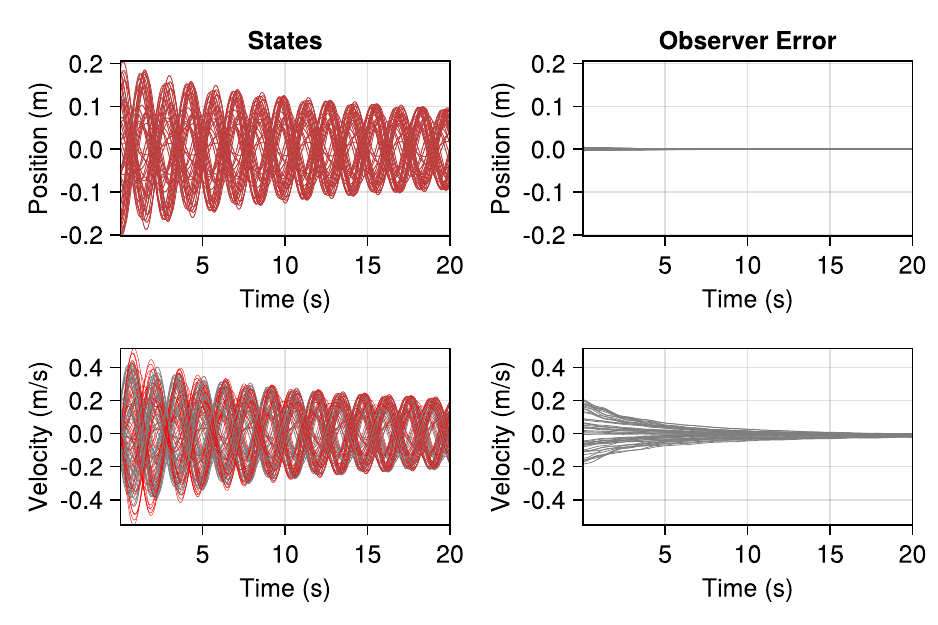}
    \caption{Simulation results showing the observer predictions and observer error with the box starting at 50 different initial conditions. The left panels compare the true (grey) and estimated (red) states, while the right panels show the observer error $x - \hat{x}$ over time. The observer error converges for all 50 test cases.}
    \label{fig:observer-results}
\end{figure}

The results are plotted in Figure \ref{fig:observer-results}. In the left-hand panels, the observer predictions (red) almost exactly match the true states (grey) after approximately 4\,s. This is confirmed by the right-hand panels, which show the observer error $x_t - \hat{x}_t$ smoothly converging to zero as the observer estimates the correct states for all simulations. 

It is worth noting that at no point did we directly train the REN to minimise the observer error. This is a natural result of using a model that is guaranteed to be contracting, and training it to minimise the one-step-ahead prediction error. There is still some residual observer error in the velocity in Figure \ref{fig:observer-results}, since our observer was only trained to approximately satisfy the correctness condition. However, this could easily be reduced or eliminated using a larger observer model, more training data, and more training epochs.

\section{Summary and conclusions} \label{sec:conc}
This paper has presented \verb|RobustNeuralNetworks.jl|, a Julia package for robust machine learning based on the recently-proposed Recurrent Equilibrium Network (REN) and Lipschitz-Bounded Deep Network (LBDN) model classes. The models are unique in that they naturally satisfy a set of \textit{built-in} robustness metrics, such as contraction and Lipschitz bounds. We have presented an overview of the model architectures, including background theory on robustness metrics in nonlinear systems, and have outlined the package structure and its usage alongside Julia's main machine-learning library, \verb|Flux.jl|. We have demonstrated via examples in image classification, reinforcement learning, and observer design that the package is easy to use in many common machine learning and data-driven control problems, while also offering the advantage of robustness guarantees.

We intend \verb|RobustNeuralNetworks.jl| to be widely-applicable in the scientific and machine learning communities for learning-based problems in which robustness certificates are crucial, and have already used the package in our own research in robust reinforcement learning \cite{Barbara++2023}. Some areas in which this package will be most applicable include: data-driven control and state estimation, image classification and segmentation, and privacy and security. We intend to expand the package with more robust neural network architectures in the future. Examples include LBDNs with one-dimensional convolution \cite{Pauli++2022c} and circular convolutions \cite{Wang+Manchester2023}, continuous-time REN models \cite{Martinelli++2023}, and RENs respecting other non-Euclidean contraction metrics \cite{Davydov++2022}. We encourage any and all contributions to \verb|RobustNeuralNetworks.jl| to further its use in robust machine learning problems.

% Acknowledgements
\paragraph*{Acknowledgements}
This work was supported in part by the Australian Research Council (DP190102963).

% **************GENERATED FILE, DO NOT EDIT**************

\bibliographystyle{juliacon}
\bibliography{ref.bib}

\end{document}